\documentclass[10pt,letterpaper]{article}
\usepackage[top=0.85in,left=2.75in,footskip=0.75in]{geometry}

\usepackage{amsmath,scalerel,amssymb}

\DeclareMathOperator*{\concat}{\scalerel*{\Vert}{\sum}}

\usepackage{algorithmic}
\usepackage{array}
\usepackage{stfloats}
\usepackage{setspace}
\usepackage{url}
\usepackage{makecell}

\usepackage{changepage}

\usepackage[utf8x]{inputenc}

\usepackage{textcomp,marvosym}

\usepackage{cite}

\usepackage{nameref,hyperref}

\usepackage[right]{lineno}

\usepackage{microtype}
\DisableLigatures[f]{encoding = *, family = * }

\usepackage[table]{xcolor}

\usepackage{booktabs}

\usepackage{array}

\newcolumntype{+}{!{\vrule width 2pt}}

\newlength\savedwidth

\raggedright
\usepackage[aboveskip=1pt,labelfont=bf,labelsep=period,justification=raggedright,singlelinecheck=off]{caption}

\makeatletter
\renewcommand{\@biblabel}[1]{\quad#1.}
\makeatother

\usepackage{lastpage,fancyhdr,graphicx}
\usepackage{epstopdf}
\fancyheadoffset[L]{2.25in}
\fancyfootoffset[L]{2.25in}
\begin{document}
\vspace*{0.2in}

\begin{flushleft}
{\Large
\textbf\newline{CAManim: Animating end-to-end network activation maps} 
}
\newline
\\
Emily Kaczmarek\textsuperscript{1,*},
Olivier X. Miguel\textsuperscript{2},
Alexa C. Bowie\textsuperscript{2},
Robin Ducharme\textsuperscript{2},
Alysha L.J. Dingwall-Harvey\textsuperscript{1,2},
Steven Hawken\textsuperscript{1,2,3,4},
Christine M. Armour\textsuperscript{1,5,6},
Mark C. Walker\textsuperscript{1,2,3,4,7,8,9,10},
Kevin Dick\textsuperscript{1,6,9*}
\\
\bigskip
\textbf{1} Children’s Hospital of Eastern Ontario Research Institute, Ottawa, Canada
\\
\textbf{2} Clinical Epidemiology Program, Ottawa Hospital Research Institute, Ottawa, Canada
\\
\textbf{3} School of Epidemiology and Public Health, University of Ottawa, Ottawa, Canada
\\
\textbf{4} ICES, Toronto, Canada
\\
\textbf{5} Department of Pediatrics, University of Ottawa, Ottawa, Canada
\\
\textbf{6} Prenatal Screening Ontario, Better Outcomes Registry \& Network, Ottawa, Canada
\\
\textbf{7} Department of Obstetrics and Gynecology, University of Ottawa, Ottawa, Canada
\\
\textbf{8} International and Global Health Office, University of Ottawa, Ottawa, Canada
\\
\textbf{9} BORN Ontario, Children’s Hospital of Eastern Ontario, Ottawa, Canada
\\
\textbf{10} Department of Obstetrics, Gynecology \& Newborn Care, The Ottawa Hospital, Ottawa, Canada
\\

\bigskip

\textcurrency Current Address: Children's Hospital of Eastern Ontario Research Institute, Ottawa, Ontario, Canada 
* \texttt{\{ekaczmarek, kdick\}@cheo.on.ca}

\end{flushleft}

\section*{Abstract}
Deep neural networks have been widely adopted in numerous domains due to their high performance and accessibility to developers and application-specific end-users. Fundamental to image-based applications is the development of Convolutional Neural Networks (CNNs), which possess the ability to automatically extract features from data. However, comprehending these complex models and their learned representations, which typically comprise millions of parameters and numerous layers, remains a challenge for both developers and end-users. This challenge arises due to the absence of interpretable and transparent tools to make sense of black-box models. There exists a growing body of Explainable Artificial Intelligence (XAI) literature, including a collection of methods denoted Class Activation Maps (CAMs), that seek to demystify what representations the model learns from the data, how it informs a given prediction, and why it, at times, performs poorly in certain tasks. We propose a novel XAI visualization method denoted CAManim that seeks to simultaneously broaden and focus end-user understanding of CNN predictions by animating the CAM-based network activation maps through all layers, effectively depicting from end-to-end how a model progressively arrives at the final layer activation. Herein, we demonstrate that CAManim works with any CAM-based method and various CNN architectures. Beyond qualitative model assessments, we additionally propose a novel quantitative assessment that expands upon the Remove and Debias (ROAD) metric, pairing the qualitative end-to-end network visual explanations assessment with our novel quantitative ``yellow brick ROAD" assessment (ybROAD). This builds upon prior research to address the increasing demand for interpretable, robust, and transparent model assessment methodology, ultimately improving an end-user's trust in a given model's predictions. Examples and source code can be found at: \url{https://omni-ml.github.io/pytorch-grad-cam-anim/}


\section*{Introduction}

The popularization of deep learning in numerous domains of research has led to the rapid adoption of these methodologies in disparate fields of scientific research. Convolutional Neural Networks (CNNs) are a class of deep learning models that use convolutions to extract image features, achieving high performance in numerous computer vision applications \cite{cnnreview}. However, due to the intrinsic network structure and the complexity of features leveraged for model predictions, CNNs are, consequently, often labeled as uninterpretable or `black-box' models. Interpretability is crucial for applications in high-criticality fields such as medicine \cite{walker2022using}, where model decisions have the potential to cause excessive harm if incorrect. In order to be deployed, models must be trustworthy both in their class predictions and in the features used to make those predictions. Therefore, there is a definitive impetus to develop trustworthy explanations of model decisions.

Presently, there exists extensive literature on the use of state-of-the-art deep learning methodologies within healthcare systems and applications. Indeed, there exist entire subfields of computer science and biomedical engineering on computational medicine and medical image analysis. Notable examples from the literature include online medical pre-diagnosis systems \cite{zhou2020cnn},
3D deep learning on medical images \cite{singh20203d}, the development of medical transformers for chest x-ray diagnosis \cite{hou2021ratchet}, and an emergent trend to adopt generative methods in these high-criticality fields (e.g. GPT-3 as a data generator for medical dialogue summarization \cite{chintagunta2021medically}). With the emergence of large language models (LLMs) such as the GPT-3 and GPT-4 models developed by OpenAI and made broadly available through the ChatGPT platform, early adopters are actively promoting the transformative opportunities of these AI systems within the healthcare space \cite{cheng2023exploring} while others issue active calls for caution in their use \cite{haupt2023ai}. Fundamentally, it is paramount to develop increasingly transparent methods to assist medical practitioners in their use of, critical oversight of, and reliance upon deep learning models. 

There have been numerous methods proposed to improve the interpretability of CNNs. Zeiler and Fergus initially investigated network interpretability by using a deconvolutional network to identify pixels activated in CNN feature maps \cite{zeiler2014visualizing}. Thereafter, gradient-based methods were used to develop saliency maps indicating important image regions based on desired output class \cite{simonyan2013deep, springenberg2014striving, sundararajan2017axiomatic}. Class Activation Maps (CAMs) are a group of methods that linearly combine weighted feature activation maps from a given CNN layer \cite{zhou16,gradcam, gradcampp, Fu20, Gildenblat21, Draelos20, Jiang21, Wang20, Desai20, eigencam}. Typically, only the final layer(s) are visualized to confer trustworthiness and describe what image features are used for model predictions. However, this provides little detail on the learning process of the model. In addition, selecting the correct final layer to visualize from each CNN model is not straightforward and is often done arbitrarily. 

To better interpret how a given model evaluates a given image through each of its layers, we propose expanding these existant Explainable Artificial Intelligence (XAI) methodologies by individually visualizing and analyzing the model's layer-wise activation maps. In a natural extension of this idea, these layer-wise activation maps can be combined as individual frames of a video animating the end-to-end network activation maps; a method we propose in this article and denote CAManim. We develop local and global normalization to understand learned network features on a layer-wise (local perspective) and network-wise scale (global perspective). We experiment and quantify layer-wise performance of CAManim with numerous CNN models and CAM variations to show performance in a variety of experimental conditions, including medical applications wherein model understanding and trustworthiness is critical. 

Our contributions are as follows:

\begin{itemize}
 \item We propose CAManim, a novel visualization method that creates activation maps for each layer in a given CNN. CAManim can be applied to any existing CNN and CAM for any classification task.

 \item We introduce local and global normalization to understand important learned features at both a layer-wise and network-wise level.

 \item We perform extensive experimentation to determine the expected time and space requirements to run CAManim.

 \item We demonstrate the usefulness of CAManim across multiple CAM variations and CNN models, and in high-criticality fields.

 \item We quantitatively evaluate the performance of each CAM visualization generated per model layer with an analytical process denoted ``yellow brick ROAD" (ybROAD) that seeks to improve the understanding of how CNNs learn. This is further extended to selecting the most accurate feature map representation from all possible layers of a CNN.
\end{itemize}

\subsection*{Related Work}

The topic of explainable and trustworthy AI has been researched extensively. Lipton \textit{et al.} \cite{lipton2018mythos} described the importance for trustworthy and interpretable models, while Ribeiro \textit{et al.} \cite{ribeiro2016should} conducted human-based trials to quantify their degree of trust in classifer predictions. Computationally, numerous methods have investigated the improvement of CNN interpretation. In this section, we provide an overview of proposed methods and how CAManim addresses a gap in the current literature.

\textbf{Earliest Explainable AI Studies:} One of the earliest efforts to interpret CNNs was made by Zeiler and Fergus\cite{zeiler2014visualizing}. In this study, feature maps from convolutional layers are used as input to a deconvolutional network to identify activated pixels in the original image space. Simonyan \textit{et al.} \cite{simonyan2013deep} approached network explainability in two ways. First, they proposed class models, which are images generated through gradient ascent that maximize the score for a given class \cite{simonyan2013deep}. Next, they produced class-specific saliency maps, calculated using the gradient of the input image with respect to a given class \cite{simonyan2013deep}.

\textbf{Guided Backpropagation and Gradient-Based Methods:} Springenberg \textit{et al.} \cite{springenberg2014striving} extended Simonyan’s work to Guided Backpropagation, which excludes all negative gradients to produce improved saliency maps. Despite calculating gradients with respect to individual classes, Selvaraju \textit{et al.} showed that the visualizations produced by Guided Backpropagation are not class-discriminative (\textit{i.e.} there is little difference between images generated using different class nodes)\cite{gradcam}. Sundarajan \textit{et al.} \cite{sundararajan2017axiomatic} proposed integrated gradients, calculated through the integral of the gradient between a given image and baseline, to satisfy axioms of sensitivity and implementation invariance. FullGrad is another gradient-based method that is non-discriminative and uses the gradients of bias layers to produce saliency maps \cite{Srinivas19}.  

\textbf{Gradient-Free Methods:} While gradient-based methods are quite popular in the field of explainable AI, some studies argue that these methods produce noisy visualizations due to gradient saturation \cite{adebayo2018sanity, kindermans2019reliability}. For this reason, gradient-free methods have been investigated by a number of studies. Zhou \textit{et al.} \cite{zhou2014object} identified \textit{K} images with the highest activation at a given neuron in a convolutional layer and occluded patches of each image to determine the object detected by the neuron. Morcos \textit{et al.} \cite{morcos2018importance} used an ablation analysis to remove individual neurons or feature maps from a CNN and quantify the effect on network performance. This study demonstrated that neurons with high class selectivity (\textit{i.e.} highly activated for a single class) may indicate poor network generalizability. Zhou \textit{et al.} \cite{zhou2018revisiting} extended this work to show that ablating neurons with high class selectivity may cause large differences in individual class performance. 

\textbf{Class Activation Maps:} A popular class of CNN visualizations are Class Activation Maps (CAMs), which produce explainable visualizations through a linearly weighted sum of feature maps at a given CNN layer \cite{zhou16}. The original CAM was proposed for a specific CNN model, consisting of convolutional, global average pooling, and dense layers at the end of the network \cite{zhou16}. The dense layer weights were used to determine the weighted importance of individual feature maps \cite{zhou16}. However, this required a specific CNN architecture and was not applicable to numerous high-performing models. This led to the development of CNN model-agnostic CAM methods.  

Gradient-based methods were the first variation of the original CAM \cite{gradcam, gradcampp, Fu20, Gildenblat21, Draelos20, Jiang21}. These methods determine importance weights by calculating averaged or element-wise gradients of the output of a class with respect to the feature maps at the desired layer. As discussed previously, gradient methods may produce noisy visualizations due to gradient saturation \cite{adebayo2018sanity, kindermans2019reliability, Wang20, Desai20, eigencam}; as a result, perturbation CAM methods have been proposed \cite{Wang20, Desai20}. In this case, importance weights are calculated by perturbing the original input image by the feature maps and measuring the change in prediction score. In addition, non-discriminative approaches have been investigated to eliminate the reliance of class-discriminative methods upon correct class predictions. For example, EigenCAM produces its CAM visualization using the principal components of the activation maps at the desired layer \cite{eigencam}.

\begin{figure*}
 \centering
 \includegraphics[width=\textwidth]{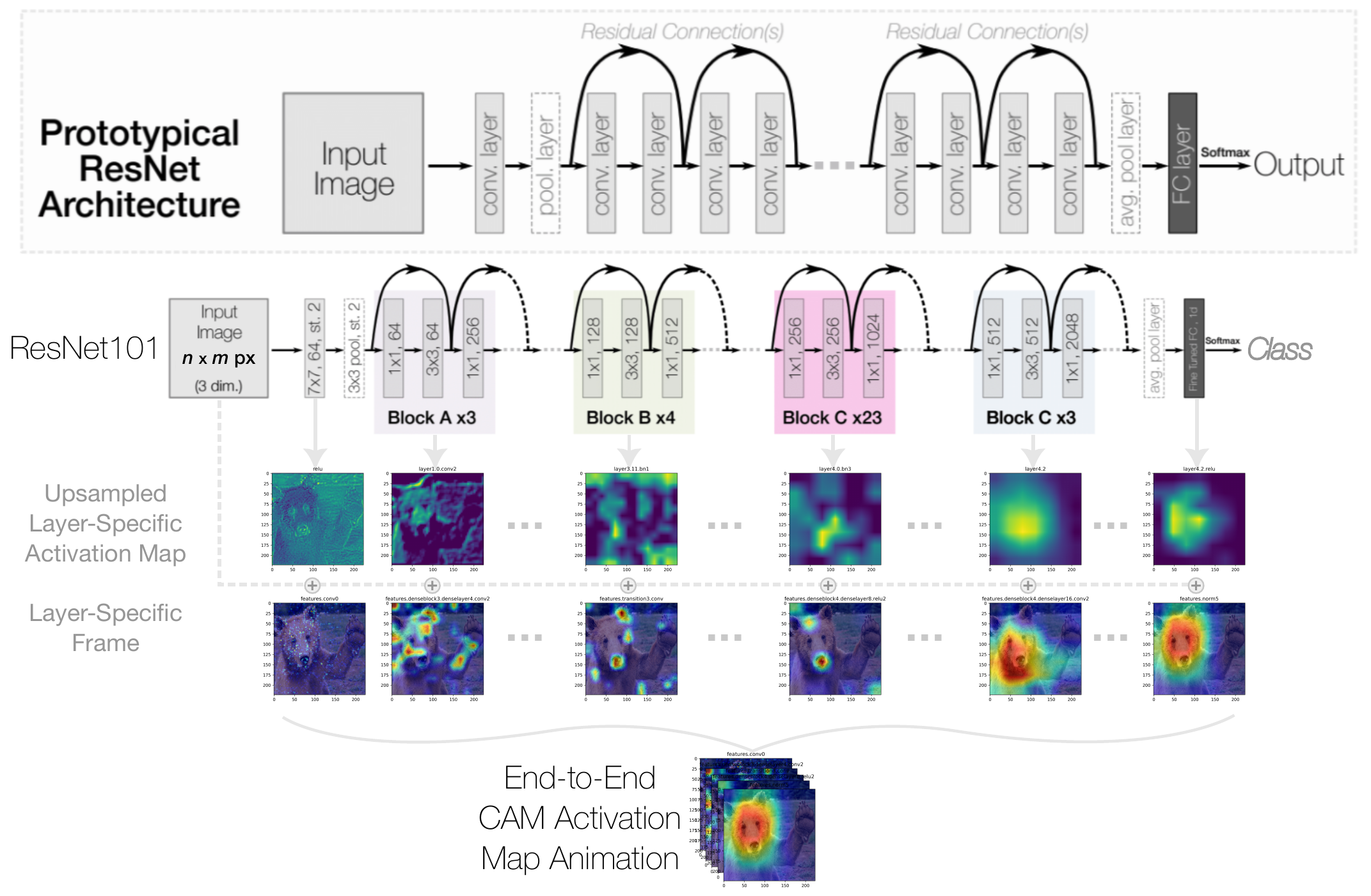}
 \caption{Conceptual overview of generating an animation of a Resnet's End-to-End Activation Maps for a given image and target class.}
 \label{fig:overview}
\end{figure*}

While most studies have developed saliency map and/or CAM formulations for a single layer, LayerCAM demonstrated how aggregating feature maps from multiple layers can refine the final CAM visualization to include more fine-detailed information \cite{Jiang21,kaczmarek2023metacam}. Gildenblat extended this idea across existing multiple CAM and saliency map methods \cite{Gildenblat21}. While conceptually similar, to the best of our knowledge, our study is the first to consider individual feature maps generated from every CNN layer and combine them into an end-to-end network explanation. Moreover, this end-to-end layer-wise analysis enables a unique view of local and global perspectives and a natural integration of both qualitative and quantitative network-wide explainability. Figure \ref{fig:overview} provides a conceptual overview of the CAManim method proposed in this work.

\section*{Materials and methods}

In this section, we first recall the general formulation for Class Activation Maps and outline notation preliminaries. Next, we explain the generation of CAManim using CAM visualizations from each layer of a CNN, depicted in Figure \ref{fig:overview}. The concepts of global and local normalization are introduced, and the computational requirements of CAManim are described from large-scale experiments. Lastly, we define the quantitative performance metric for individual CAM visualizations, and propose ybROAD for analyzing end-to-end layer-wise CAManim. 

\subsection*{Individual CAM Formulation}

The general formulation for any CAM method consists of taking a linearly weighted sum of feature maps as follows: 

\begin{equation}\label{eqn:cam-form1}
L^c_{CAM(A^l)} = \sum\limits_{k}(\alpha_k^cA_k^l),  \textit{where }  A^l= f ^l(x)
\end{equation}

For a given input image $x$ and CNN model $f(\cdot)$, a CAM visualization $L$ can be generated through the weighted \(\alpha\) summation of $k$ activation feature maps $A$ at layer $l$. Class discriminative CAM methods further define $L$ per predicted class $c$. To exclude negative activations, most CAM formulations are followed by a ReLU operation.

\subsection*{End-to-End Layerwise Activation Maps}

To formulate CAManim, CAM visualizations are first generated for every differentiable layer $l$ within a given CNN with a total number of layers $N$:

\begin{equation}\label{eqn:cam-form2}
L^c_{CAManim} = L^c_{CAM(A^{l=0})}, \ldots , L^c_{CAM(A^{l=N})}
\end{equation}

Each CAM visualization is subsequently saved as a PNG image $I$ and concatenated together to create the final CAManim video, as depicted below:

\begin{equation}\label{eqn:cam-form3}
CAManim = \concat_{l}^{N}I_{L^c_{CAM(A^{l})}} 
\end{equation}

For clarity, the concatenation operator, $\concat$, is defined in this work in a way analagous to the summation operator, $\Sigma$, and product operator, $\Pi$, to concisely express the sequential organization of individual frames into the resulting animated video.

\subsection*{Global- vs. Local-Level Normalization}
For a model end-user to correctly interpret what importance the model attributes to particular pixels at a given layer in the network, they must be provided the appropriate context. To this end, the model interpreter may wish to know ``\textit{what importance does the model place on particular pixels at a given layer?}" or ``\textit{what importance does the model place upon particular pixels overall?}". Consequently, two normalization approaches can be leveraged, each with the intent of correctly relaying information to answer one of these two questions, and both complementary to the other. Thus CAManim visualizes the CAM activations of each layer using two different types of normalization: Local-Level Normalization and Global-Level Normalization. 
Global normalization is performed using the minimum and maximum activation value across all activations generated, which is practical for determining and visualizing which layer contains the strongest network-wide activation for a given class. Local normalization uses the minimum and maximum values of the activations of each specific layer. Local normalization, contrary to global-level normalization, depicts the strongest layer-level activation  and therefore provides layer-wise information.

\begin{figure}
 \centering
 \includegraphics[width=\textwidth]{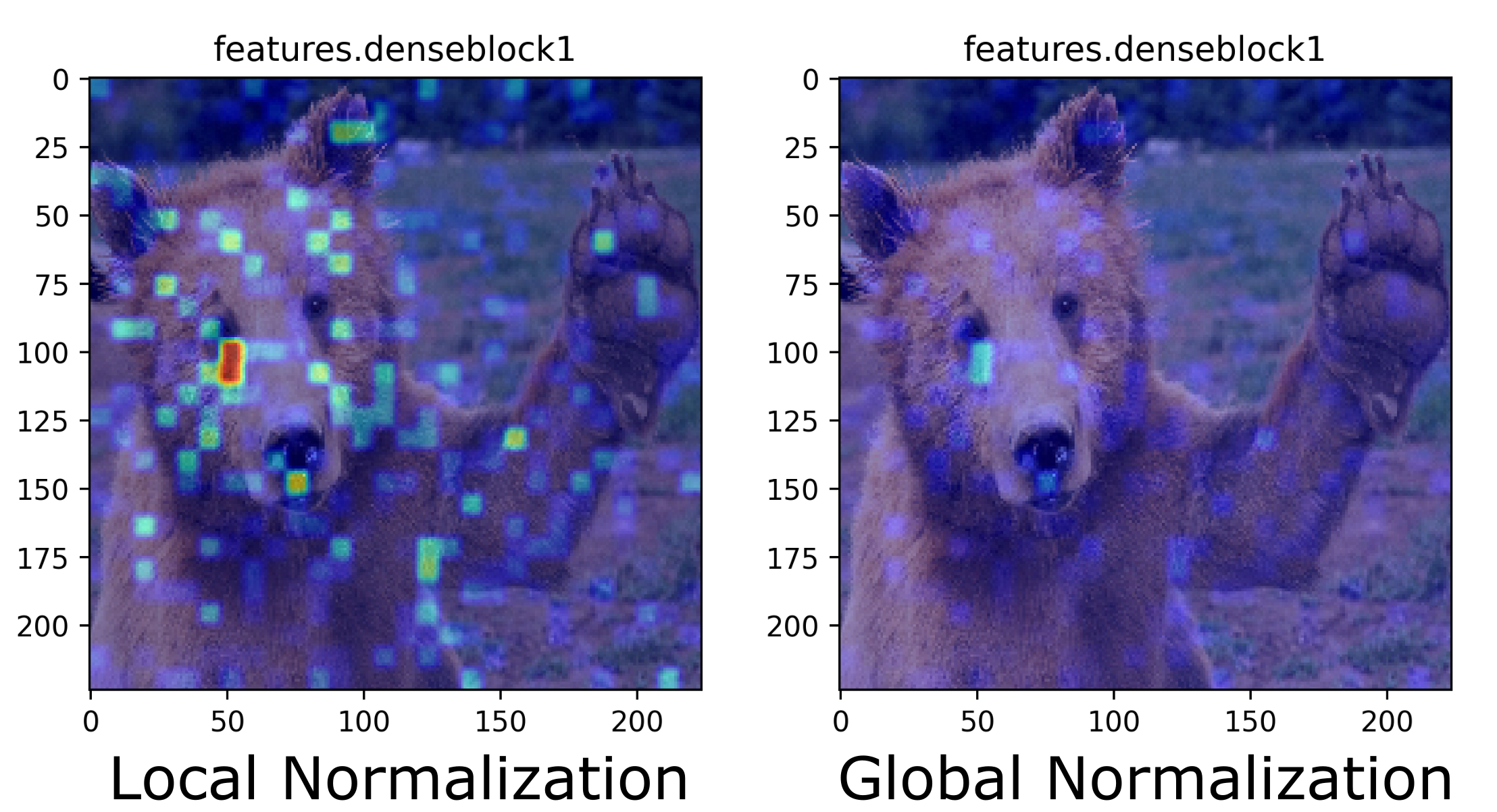}
 \caption{Difference between local and global normalization for the feature map generated from layer \textit{features.denseblock1} in DenseNet161.}
 \label{fig:normalization}
\end{figure}

Figure \ref{fig:normalization} shows the difference between global and local normalization for the first denseblock of DenseNet161 \cite{Huang16}. The global normalization (right) displays an attenuated version of the local normalization (left). This example demonstrates that the layer-wise information focuses upon learning small pattern-like features, whereas the network-wise information indicates that the activations of this layer are generally attenuated with respect to all other layers within the DenseNet161 model.

\subsection*{Model- \& CAM-Specific Interpretation and Computational Complexity}

To appreciate how varying CNN architectures and CAM methods produce differing visual explanations for a given image $x$, target class $c$, and CNN model $f(\cdot)$, we ran large-scale experiments producing numerous layer-wise and globally normalized CAManim videos/image sequences. Consequently, this additionally enabled the benchmarking of key model-specific metrics such as layer-level number of parameters and CAManim run-time.

\begin{figure}
 \centering
 \includegraphics[width=\textwidth]{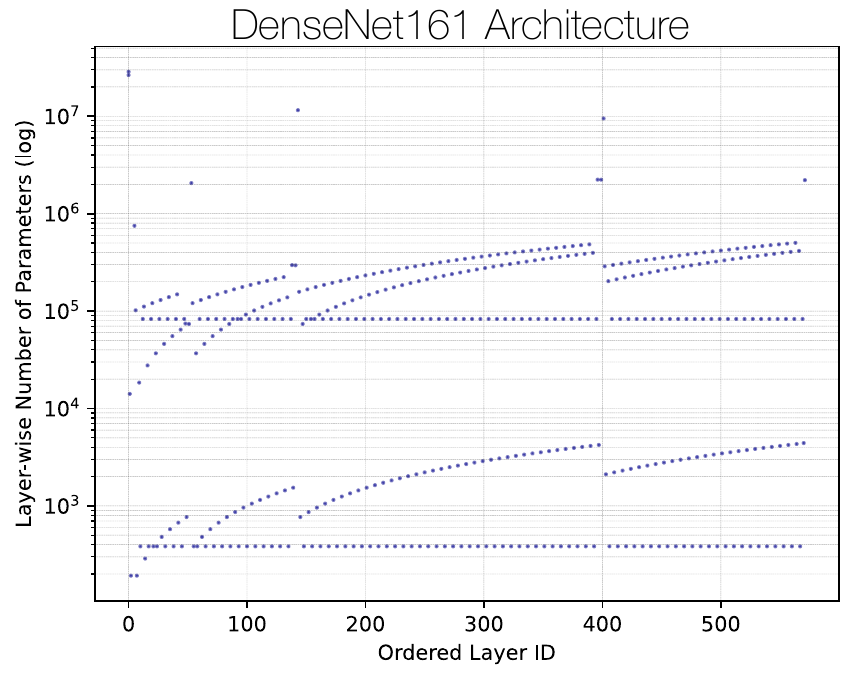}
 \caption{Layer-wise Depiction of DenseNet161 parameters.}
 \label{fig:densenet-params}
\end{figure}

Figure \ref{fig:densenet-params} illustrates the layer-wise parameter number along a log-scale where we can explicitly visualize the four general \texttt{DenseBlocks} comprised of a varying number of \texttt{DenseLayers}. Since CAManim computation will vary by layer number $n$, layer-wise parameters $p$, and CAM-specific compute runtime $r$, we generally estimate that CAManim will have a simplified computational time complexity of $\mathcal{O}(n\bar{p}\bar{r})$. For clarity, the overbar notion expresses averages for the number of parameters and CAM-specific compute time, respectively. Image-specific dimension will also impact runtime, however, given that the majority of models reshape their input to a consistent size, this constant factor may be subsumed within term $n$. To provide general estimates on the overall runtime for a given CNN and CAM, we tabulate in Table \ref{tab:ests} our experimental benchmarks using an Intel Xeon CPU \@2.20 GHz, 13GB RAM, Tesla K80 GPU accelerator, and 12GB GDDR5 VRAM.

\begin{table}[b]
\caption{Total number of parameters, CAManim runtime, and average parameters and runtime across all layers included in CAManim calculated for six CNN models using HiResCAM}
\scriptsize
\centering
\begin{tabular}{ rcccc } 
 \toprule
 \textbf{Model} & Num. Params. & Time (s) & \makecell{Avg. Params.\\ per Layer} & \makecell{Avg. Layer \\Time (s)} \\ \midrule
\textbf{AlexNet} & 61,100,840 & 16.67 & 329,292.8 & 0.2622 \\ 
 \textbf{ConvNeXT} & 88,591,464 & 392.78  & 1,140,535.1 & 0.2665 \\ 
 \textbf{DenseNet161} & 28,681,000 & 514.99 & 180,814.6 & 0.0997 \\ 
 \textbf{EfficientNet-b7} & 66,347,960 & 972.42  & 345,325.1 & 0.1375 \\
 M\textbf{axViT-t} & 30,919,624 & 596.84  & 292,907.1 & 0.1080 \\ 
 \textbf{SqueezeNet} & 1,248,424 & 52.54  & 48,030.8 & 0.0144 \\ 
 \bottomrule
\end{tabular}
\label{tab:ests}
\end{table}

\subsection*{Quantitative Evaluation}
To quantitatively evaluate the performance of each CAM visualization and demonstrate the information gained through deeper layers in a CNN, we calculate the Remove and Debias (ROAD) score \cite{Rong22}. This metric has superior computational efficiency and prevents data leakage found with other CAM performance metrics \cite{Rong22}. ROAD perturbs images through noisy linear imputations, blurring regions of the image based on neighbouring pixel values \cite{Rong22}. The confidence increase or decrease \textit{C} in classification score using the perturbed image with the \textit{least relevant pixels} (LRP) or \textit{most relevant pixels} (MRP) is then used to evaluate the accuracy of a CAM visualization. Since the percentage of pixels perturbed affects the ROAD performance, we evaluate ROAD at \textit{p =} 20\%, 40\%, 60\% and 80\% pixel perturbation thresholds. As proposed by Gildenblat \cite{Gildenblat21}, we combine the LRP and MRP scores for our final metric: 

\begin{equation}
ROAD(L^c_{CAM(A^{l})}) = \sum\limits_{p} \frac{(C^p_{L_{LRP}} - C^p_{L_{MRP}})}{2}
\end{equation}

A ROAD score is calculated for each CAM generated. Therefore, for \textit{N} differentiable layers in a CNN, there will be \textit{N} ROAD scores calculated within CAManim. Given that this network-wide sequence of ROAD values represents a journey-like traversal of the network, we denote this series of ROAD values as the 'yellow brick ROAD', or ybROAD for brevity:

\begin{equation}\label{eqn:cam-form4}
ybROAD = \concat_{l}^{N}ROAD(L^c_{CAM(A^{l})})
\end{equation}

The ybROAD scores can be used to analyze performance of an experiment with given image $x$, target class $c$, and CNN model $f(\cdot)$ over all layers of the network. Consequently, this analysis enables the quantitative identification of the CNN layer that maximally visualizes features with the largest impact on model performance through \texttt{max}(ybROAD). The \texttt{mean}(ybROAD) score is also calculated to summarize the overall model end-to-end ROAD performance. Interestingly, variant metrics derived from ybROAD values may yield new insights into the quantification of a model's ability to predict particular classes.

\section*{Results \& Discussion}
In this section, we first define the pre-trained models and datasets used to evaluate CAManim. Next, we demonstrate CAManim in high-criticality fields using a ResNet50 model fine-tuned to perform breast cancer classification. We then show example visualizations from CAManim for 10 different CAM variations and discuss abnormal visualizations. Lastly, we discuss the ybROAD performance of CAManim and future directions building upon this work.

\subsection*{Pre-trained Models and Datasets}
\label{sec:datamodel}
To evaluate CAManim, we use models from Pytorch pre-trained on the 2012 ImageNet-1K dataset \cite{deng2009imagenet}. Specifically, results are shown for AlexNet \cite{alexnet12}, ConvNeXT \cite{convnext22}, DenseNet161 \cite{Huang16}, EfficientNet-b7 \cite{efficientnet19}, MaxViT-t \cite{tu2022maxvit}, and SqueezeNet \cite{SqueezeNet}. The CAManim videos for an additional 14 models and publicly available code can be found here: \url{https://omni-ml.github.io/pytorch-grad-cam-anim/}. All results in this study (apart from the high-criticality case study) leverage a popular brown bear-containing image typically used in the CAM research community; following an emergent standard, the image is preprocessed by resizing to 224 $\times$ 224 and normalized. Next, we demonstrate the utility of CAManim in a high-criticality field. 

\subsection*{Case Study: End-to-End BC-ResNet50 Visualization for Malignant Tumour Prediction}
\label{sec:bcresnet}
We leverage a ResNet50 model \cite{He15} initially trained on the 2012 ImageNet-1K dataset \cite{deng2009imagenet} and fine-tune the model using the Kaggle breast ultrasound data to classify malignant vs. normal images \cite{kagglebc}. For simplicity, we refer to this network as BC-ResNet50 (``Breast Cancer-ResNet50"). This dataset comprises 133 normal images and 210 malignant images that are split into a 80-10-10\% train-validation-test split. Images are preprocessed to a size of 224 $\times$ 224 and various augmentations are applied to the training set. 

\begin{figure*}
 \centering
 \includegraphics[width=\textwidth]{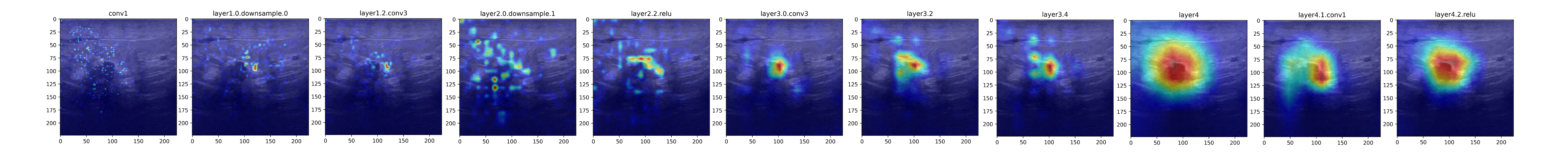}
 \caption{Visualization of the activation maps from BC-ResNet50 to visually depict how the model predicts the 'malignant' tumour class. Only the 10th percentile layers are illustrated for concision.}
 \label{fig:mednet-viz}
\end{figure*}

Pre-processing and training steps are selected based on MONAI recommendations \footnote{\url{https://github.com/Project-MONAI/tutorials/blob/main/2d_classification/mednist_tutorial.ipynb}}. Following fine-tuning, CAManim is run with an example test image of the malignant class to visualize and interpret how the resultant CNN arrives at producing the correct prediction of malignant cancer. Figure \ref{fig:mednet-viz} illustrates the layer-wise activations that BC-ResNet50 considers when determining the `malignant' tumour.

It is important to emphasize that for high-criticality applications such as medical imagery, the initial resizing of input imagery can dramatically alter the information available to the model and impact model out and its interpretability. While this work builds upon previous XAI literature and adopts their methodological approach, we recommend that for high-criticality applications, the initial image size be kept closely aligned with original input image sizes (no/limited downsizing) so as not to alter image resolution and to provide a clinical decision support system as a visual explanation method.

\subsection*{Visualizing End-to-End Network Activation Maps}
\label{sec:activationmaps}

We further demonstrate the performance of CAManim on 10 different CAM methods, including seven gradient-based methods (EigenGradCAM, GradCAM, GradCAMElementWise, GradCAM++, HiResCAm, LayerCAM, and XGradCAM), two perturbation methods (AblationCAM and ScoreCAM), a principal components method (EigenCAM), and RandomCAM. RandomCAM generates random feature activation maps from a uniform distribution between $[-1,1]$.

\begin{figure*}
 \centering
 \includegraphics[width=\textwidth]{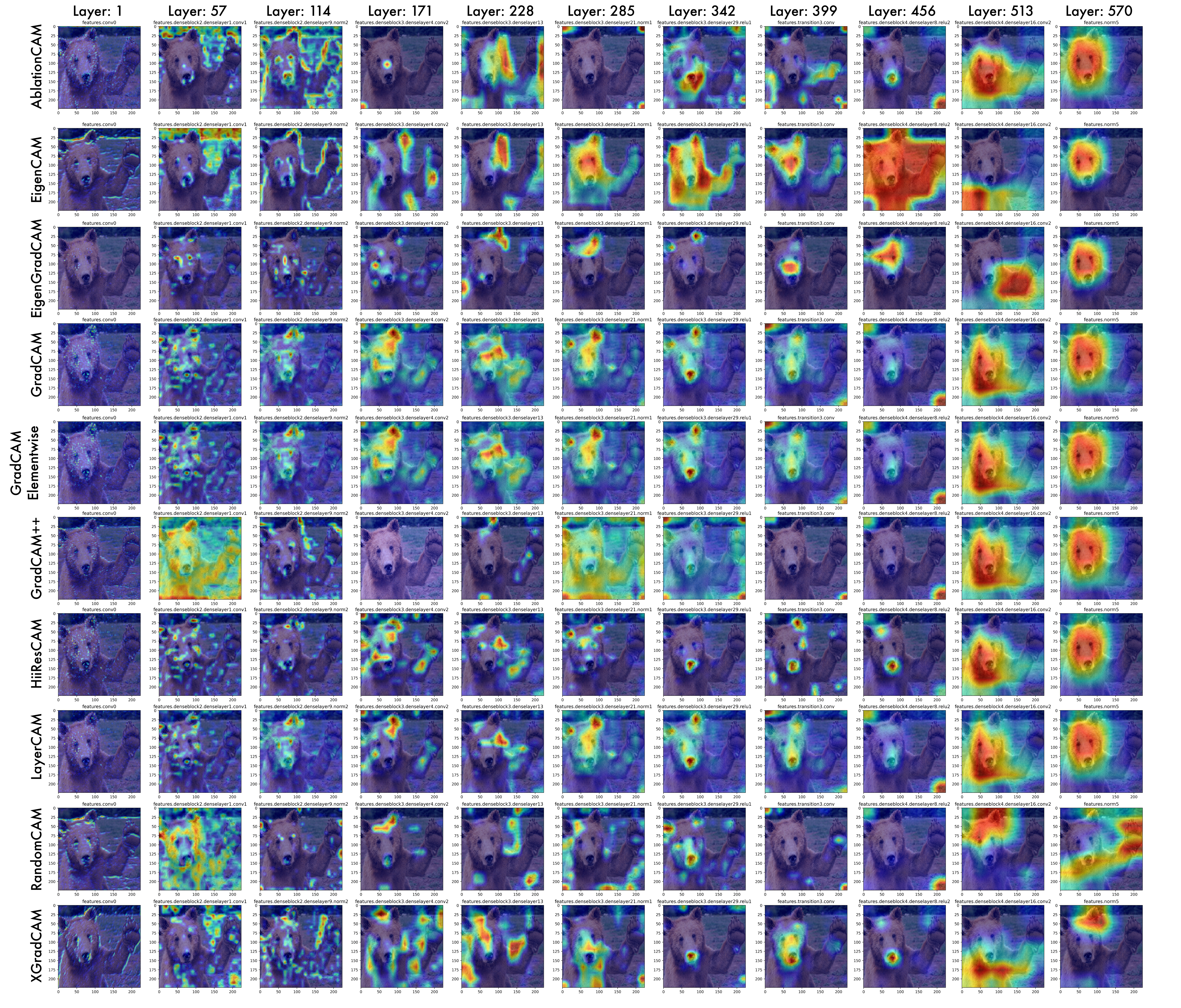}
 \caption{End-to-end activation map visualization for 10 CAMs using DenseNet161. Every 10th percentile map is depicted, from left to right.}
 \label{fig:all-cams}
\end{figure*}

\begin{figure}
 \centering
 \includegraphics[width=0.9\textwidth]{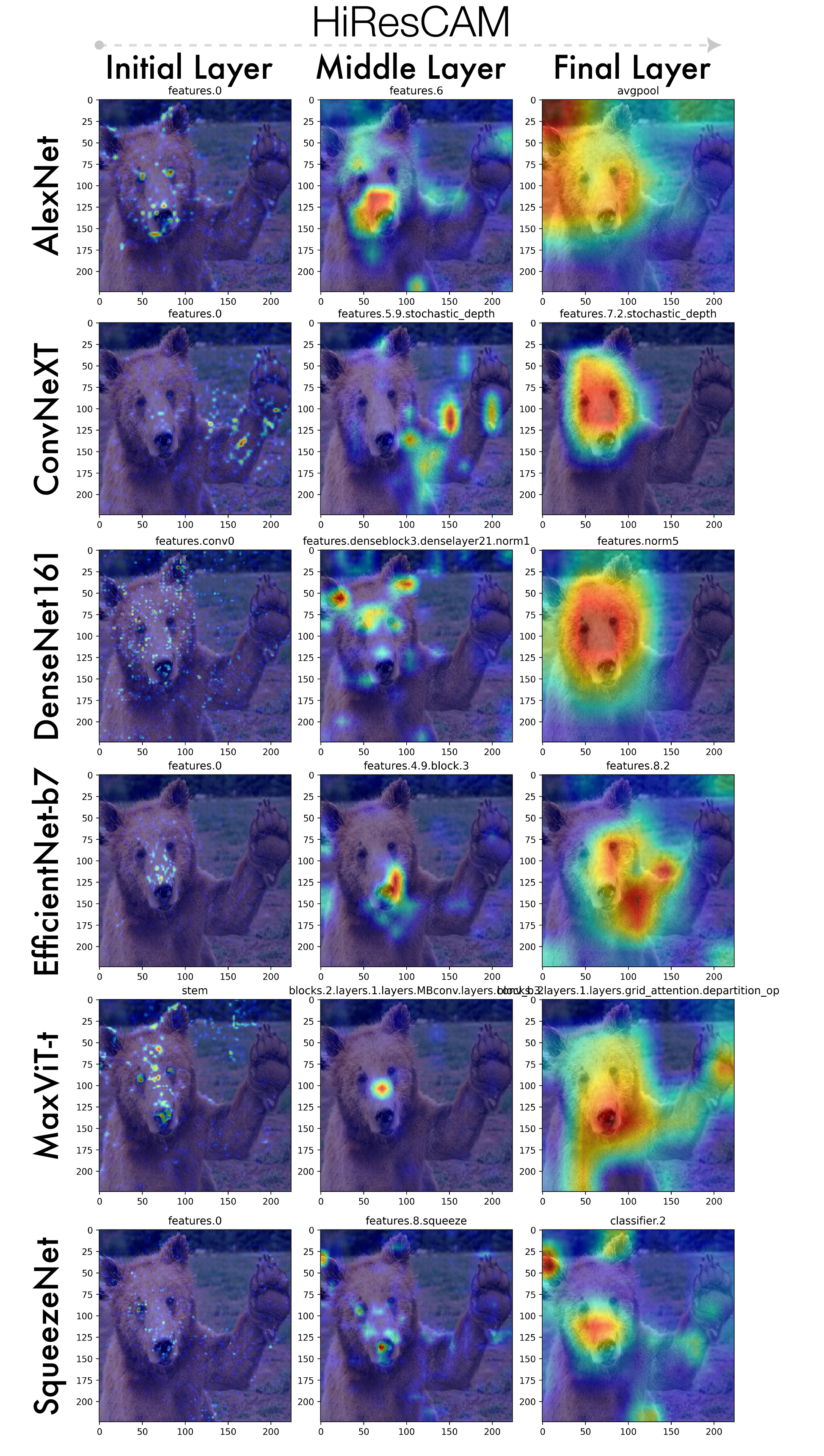}
 \caption{Initial, middle, and final activation maps applying a single CAM, HiResCAM, to various model architectures.}
 \label{fig:six-models}
\end{figure}

As expected, Figures \ref{fig:all-cams} \& \ref{fig:six-models} depicts early model layers as activating general patterns and edges while middle and final layers progressively focus the activation maps to regions highly characteristic of the brown bear contained within. Such a layer-wise approach enables the pair-wise or multi-wise comparison of visual-explanation methods and how these individual activation maps compare globally across all activation maps.

\subsection*{Layer-Type Visualization Issues}
\label{sec:visissues}

\begin{figure}
 \centering
 \includegraphics[width=\textwidth]{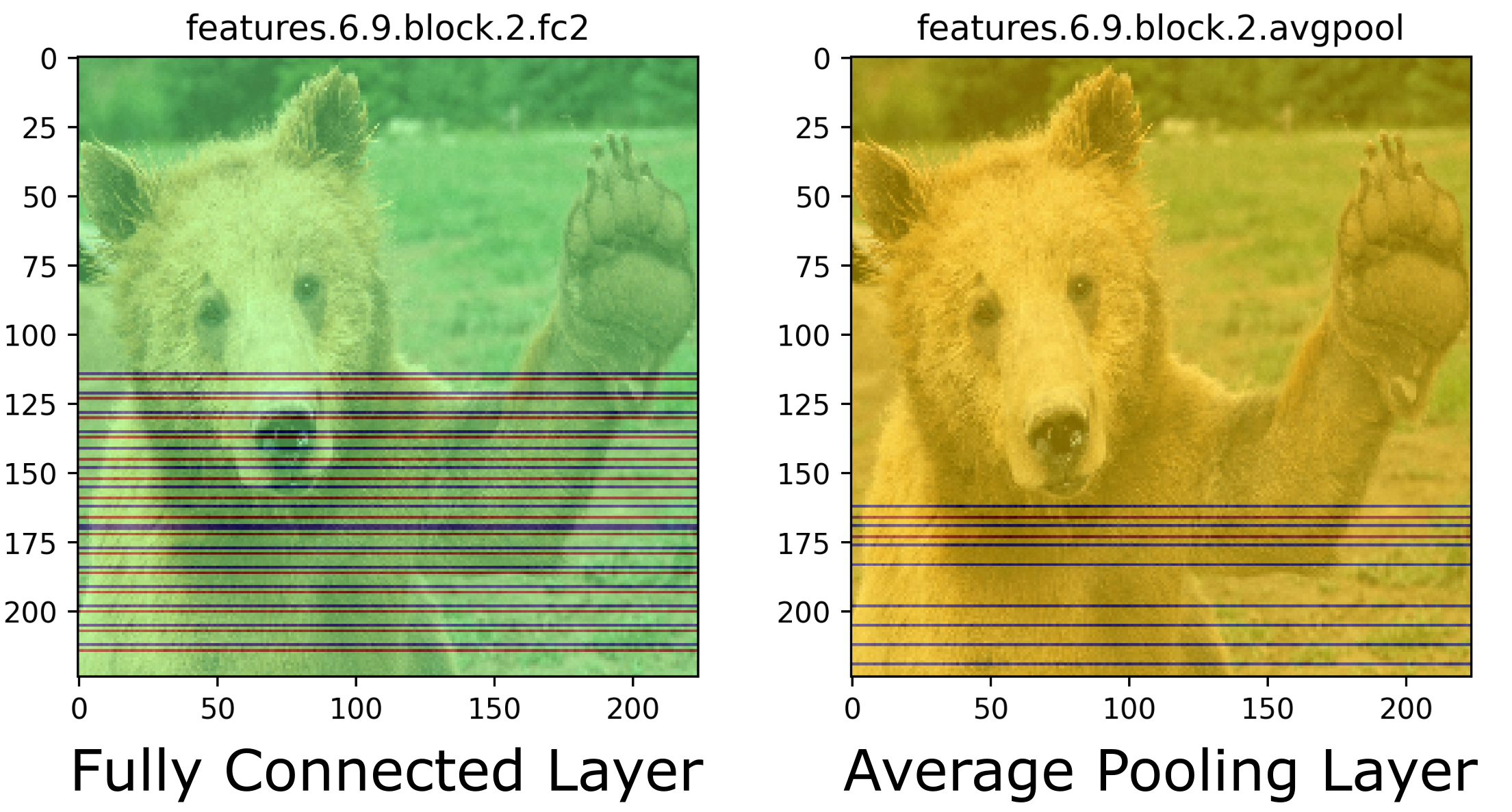}
 \caption{Visualization of CAManim for fully connected and average pooling layers.}
 \label{fig:bad_layers}
\end{figure}

Certain differentiable layers may produce unanticipated CAM visualizations, as depicted in Figure \ref{fig:bad_layers}. In these layers, images are compressed to 1-dimensional (1D) representations; consequently, 2D feature visualization of a non-convolutional layer is effectively meaningless. Instead, individual neurons that are highly activated show up as vertical or horizontal lines across the image. While these images are uninformative, they simply depict visualizations of 1D vectors and should be filtered out.

\subsection*{ybROAD Quantitative Evaluation}
\label{sec:ybROAD}

\begin{figure}
 \centering
 \includegraphics[width=\textwidth]{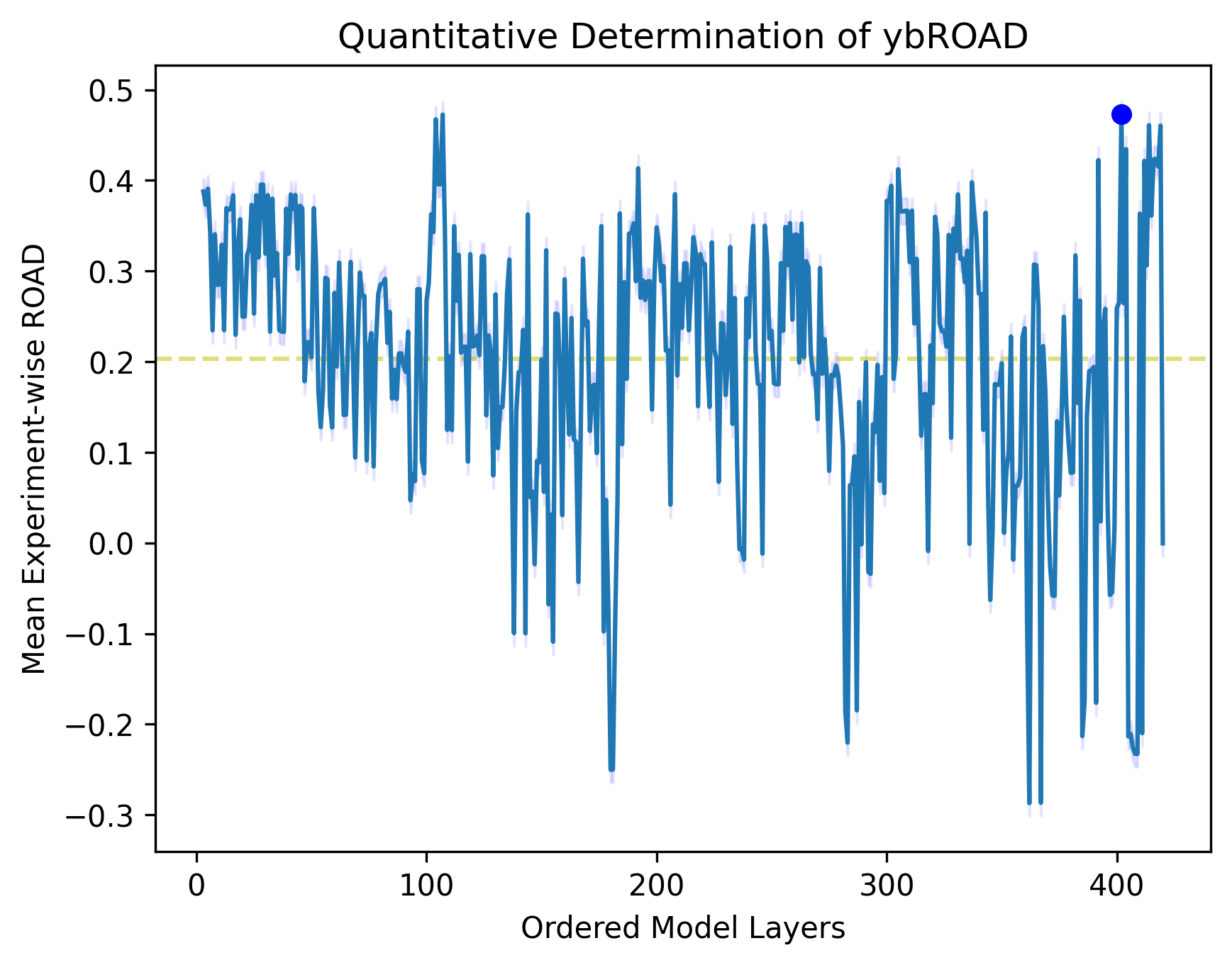}
 \caption{Depiction of end-to-end ROAD values, denoted ybROAD.}
 \label{fig:ybROAD}
 \vspace{-12px}
\end{figure}

Figure \ref{fig:ybROAD} displays the ybROAD for 11 trials of generating CAManim for the bear image using ResNet152 (mean ybROAD: 0.204; max ybROAD: 0.473 at layer 402). Initially, the layer-wise ROAD performance is very high ($\sim$0.4). At this point, the CNN layer is activating many small regions throughout the image; when each of these areas is perturbed, it is difficult to correctly classify the image, and the ROAD score increases. As the network starts learning larger features, less of the bear image is perturbed, and the ROAD score decreases. Towards the end of the network, the ROAD score increases again and reaches its maximum value as the small activations are combined together to encapsulate the entire bear. This demonstrates how the ybROAD score can provide more information on how the network progressively learns. 

\begin{figure}
    \centering
    \includegraphics[width=\textwidth]{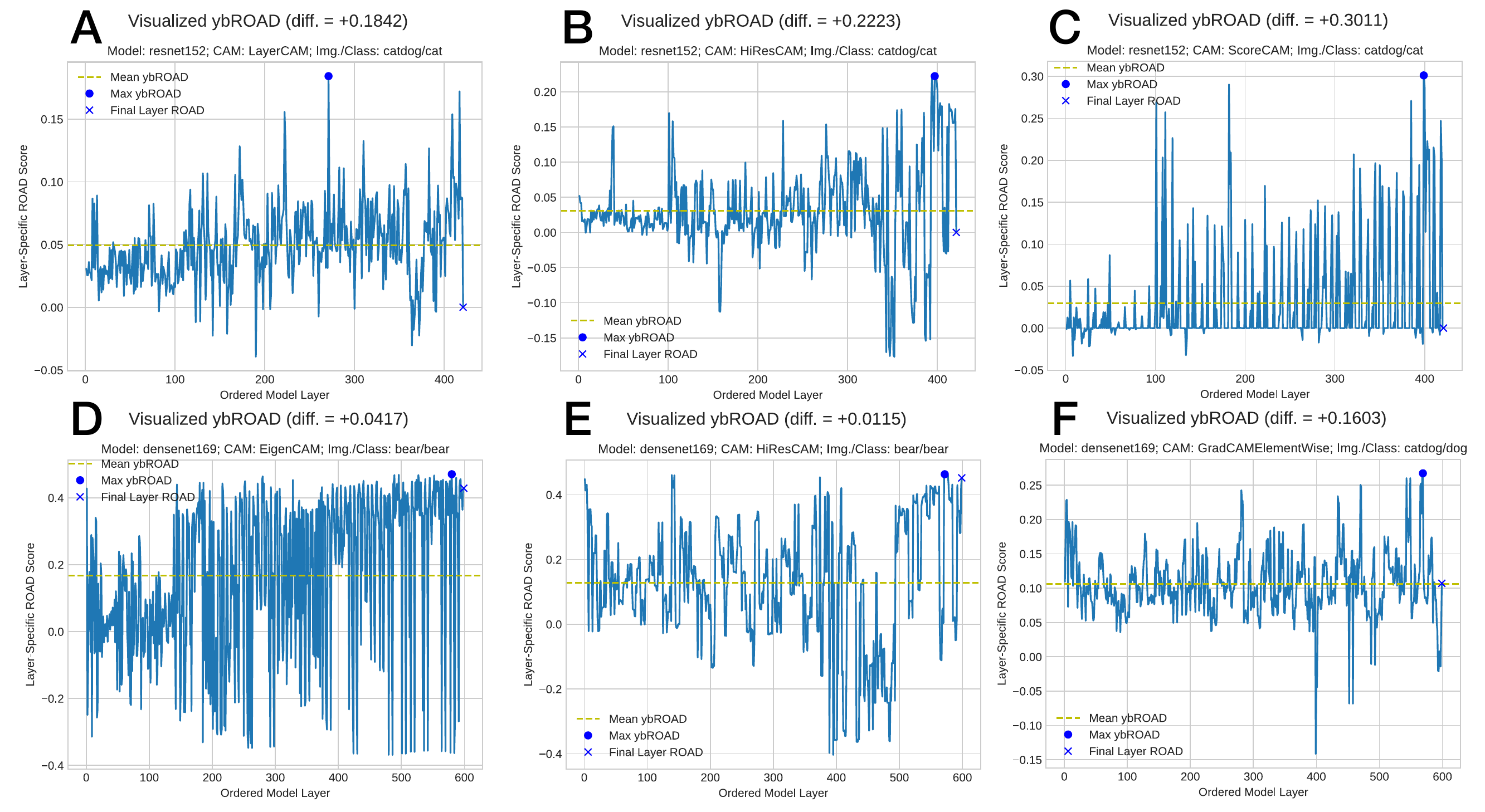}
    \caption{Quantitative determination of ybROAD \& visualization of model convergence to the target class.}
    \label{fig:yb-converge}
\end{figure}

Interestingly, the layer-wise depiction of ROAD values may be used to investigate how various model layers contribute to the overall discrimination of a given target class within an image for a pre-trained model of a given architecture and selected CAM. To quantify the improvement of our ybROAD method against standard practise (\textit{i.e.,} considering the activation map of the final layer of a model), we sumarize 12 diverse experimental conditions in Table \ref{tbl-compare}. The difference in the ybROAD vs. final layer-ROAD values is indicative of the performance improvement from our proposed layer-wise approach. Figure \ref{fig:yb-converge} additionally depicts the general improvement and convergence of ROAD values across all model layers. Interestingly, this layer-wise series of values affords greater insight into the general functioning and utility of various model layer contributions across experiments. While Figures \ref{fig:yb-converge}A,B,D,E all seem to generally improve in ROAD performance from model layer beginning to end, Figures \ref{fig:yb-converge}C,F both appear relatively consistent in their value distribution,  perhaps indicative that within these instances, the model/CAM combination had greater difficulty in discriminating the target class within the given image. Certainly, across all experiments we observe a noisy time-series signal suggesting that future work investigate moving average smoothing as a technique to make these curves more interpretable, albeit, as a trade-off for the layer-specific resolution of ROAD values.

The combined consideration of quantitative ROAD and qualitative CAM at every layer enables end-users to identify the best representation for their particular image, target class, and model in a manner less arbitrary than selecting one of several terminal layers. For example, a healthcare professional might identify a better representative feature map for a predicted tumour than they might otherwise from a potentially poorer last-layer visualization. This approach effectively allows an end-user to peer across the network and determine those layers that best capture the story as opposed to relying on the final output alone. We caution that this may introduce additional risk for confirmation bias, however, this has broadly been a challenge within the XAI community.

\begin{table}[]
\caption{\label{tbl-compare}Quantitative comparison of ybROAD values and SOTA CAM methods for various model architectures, images, and target classes.}
\resizebox{\textwidth}{!}{\begin{tabular}{cccccccc}
\toprule
\textbf{\begin{tabular}[c]{@{}c@{}}Model\\ Architecture\end{tabular}} & \textbf{\begin{tabular}[c]{@{}c@{}}Image\\ Name\end{tabular}} & \multicolumn{1}{c}{\textbf{\begin{tabular}[c]{@{}c@{}}Target\\ Class\end{tabular}}} & \textbf{\begin{tabular}[c]{@{}c@{}}Selected\\ CAM\end{tabular}} & \textbf{\begin{tabular}[c]{@{}c@{}}Mean \\ Layer-wise \\ ROAD\end{tabular}} & \textbf{\begin{tabular}[c]{@{}c@{}}Final Layer \\ ROAD\end{tabular}} & \textbf{\begin{tabular}[c]{@{}c@{}}ybROAD\end{tabular}} & \textbf{\begin{tabular}[c]{@{}c@{}}Difference\\ (ybROAD - Final)\end{tabular}} \\ \midrule
ResNet152 & \multicolumn{1}{c}{catdog} & dog  & ScoreCAM & 0.133 & -3.50E-07 & 0.499 & \cellcolor[HTML]{B1CF95}+0.499  \\
ResNet152 & bear   & bear & ScoreCAM & 0.107 & -6.26E-07 & 0.491 & \cellcolor[HTML]{B2D097}+0.491  \\
ResNet152 & bear   & bear & EigenCAM & 0.153 & -5.96E-08 & 0.486 & \cellcolor[HTML]{B3D097}+0.486  \\
ResNet152 & catdog & cat  & ScoreCAM & 0.029 & -2.92E-07 & 0.301 & \cellcolor[HTML]{C6DCB1}+0.301  \\
ResNet152 & catdog & dog  & GradCAMElementWise & 0.120 & 4.26E-05 & 0.228 & \cellcolor[HTML]{CEE1BD}+0.228  \\
ResNet152 & catdog & cat  & HiResCAM & 0.030 & 1.74E-06 & 0.222 & \cellcolor[HTML]{CEE1BE}+0.222  \\
DenseNet169 & catdog & dog  & EigenCAM & 0.043 & 0.124 & 0.334 & \cellcolor[HTML]{CFE2BF}+0.210  \\
DenseNet169 & catdog & dog  & LayerCAM & 0.110 & 0.111 & 0.309 & \cellcolor[HTML]{D1E3C1}+0.197  \\
ResNet152 & catdog & cat  & LayerCAM & 0.049 & 1.20E-04 & 0.184 & \cellcolor[HTML]{D2E4C3}+0.184  \\
DenseNet169 & catdog & dog  & GradCAMElementWise & 0.106 & 0.107 & 0.267 & \cellcolor[HTML]{D5E5C6}+0.160  \\
DenseNet169 & bear   & bear & EigenCAM & 0.167 & 0.429 & 0.470 & \cellcolor[HTML]{E2EDD8}+0.042  \\
DenseNet169 & bear   & bear & HiResCAM & 0.128 & 0.452 & 0.463 & \cellcolor[HTML]{E4EFDC}+0.011 \\ \bottomrule
\end{tabular}}
\end{table}

\subsection*{CAM Failure Cases}
\begin{figure}
 \centering
 \includegraphics[width=0.99\textwidth]{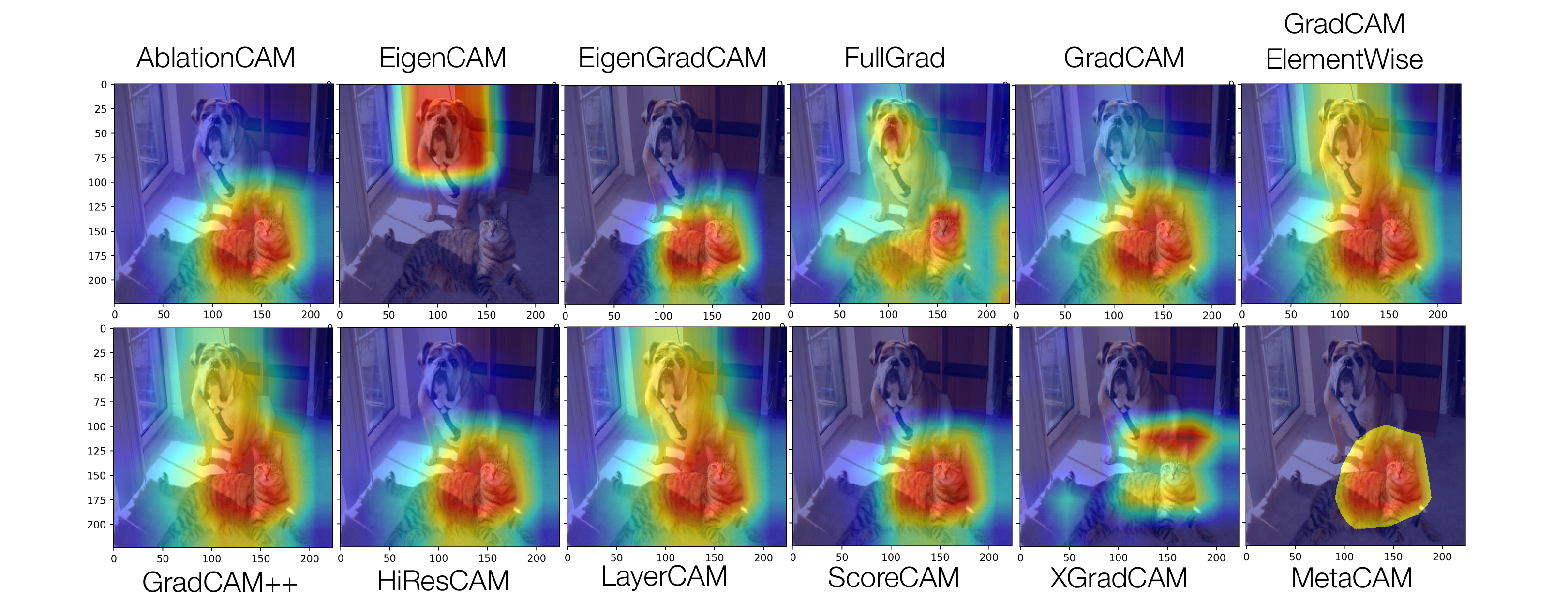}
 \caption{Multiple CAM demonstration with varied target classes (`catdog` image with target classes ) and EigenCAM failure case.}
 \label{fig:failure-cases}
\end{figure}

Interestingly, EigenCAM incorrectly highlights the dog in the image, instead of the desired cat class. This explains the negative ROAD value for EigenCAM in Figure \ref{fig:failure-cases}. EigenCAM is a non-discriminative CAM method that uses principal components to create activation maps. However, when there are multiple classes within the same image, the order of principal components must be specified (\textit{e.g.,} first principal components vs. second principal components). EigenCAM performs well on images with a ``single-subject", but otherwise requires a user to determine the number and rank of various components within an image to perform successfully. This requires a level of hand-engineering and data leakage to correctly align the appropriate principal component with the intended class.

\begin{figure}
    \centering
    \includegraphics[width=\textwidth]{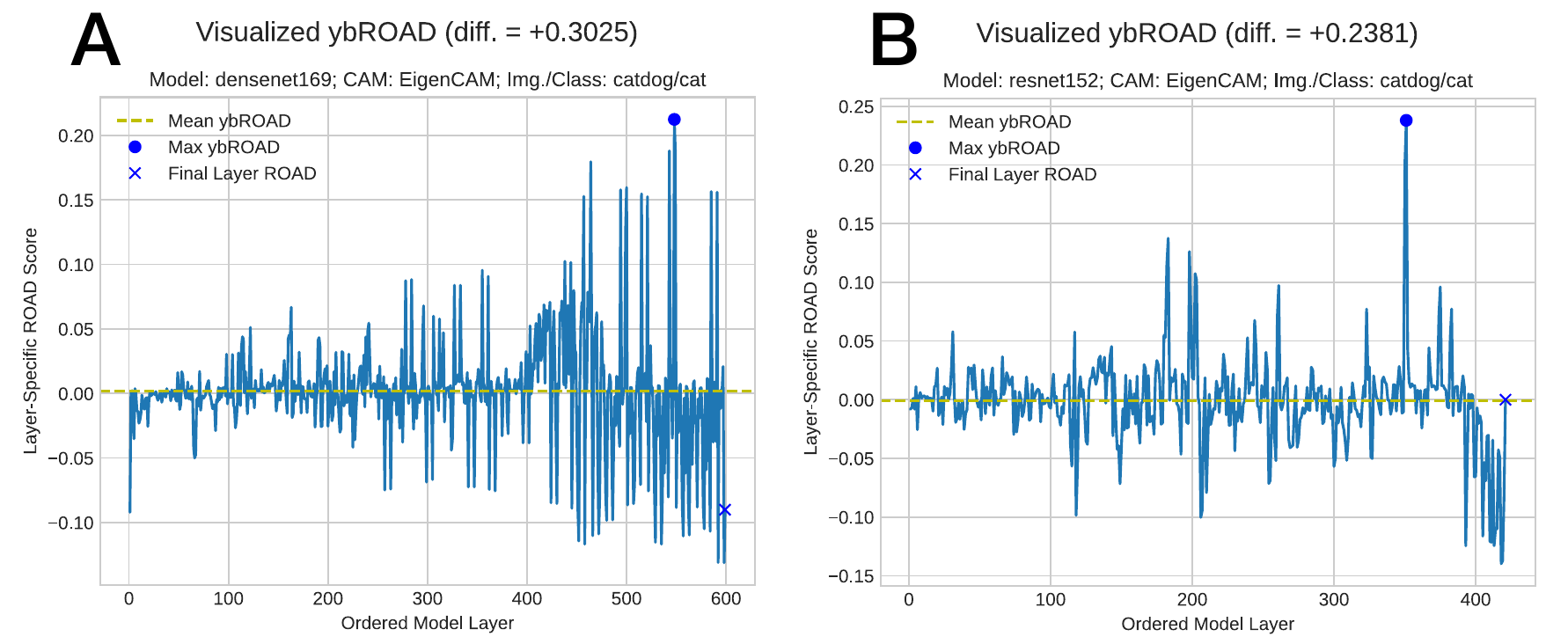}
    \caption{Failure case where the ybROAD plot indicates that the model was unable to correctly distinguish the correct object within this image, class, model, and CAM combination.}
    \label{fig:yb-fail}
\end{figure}

The ybROAD plots proposed within this work can be leveraged to better understand whether a model adequately distinguishes a given class or whether it fails across all layers of the model. As visualized in the ybROAD plots of Figure \ref{fig:yb-fail} the mean layerwise ROAD value around 0 effectively demonstrate that the model was unable to identify the correct class within the image. Consequently, the ybROAD quantitative metrics derived from the ybROAD plots may be useful in elucidating the impact of model architecture (and their learned parameters) on a class-specific basis. As part of future work, this concept could be extended to consider epoch-wise ybROAD plots to better determine how specific layers through model training contribute to the discrimination of the target class.

\subsection*{Future Directions}
The proposed future directions for research represent individual contributions that can significantly advance the use of CAMs for CNNs. Foremost, conducting more in-depth studies on the activation maps statistics at different layers and for different images can provide a better understanding of how CNNs attend to images in varying applications and contexts. Secondly, designing an algorithm to efficiently compute CAM-based videos would greatly improve the applicability of this technique in various fields, particularly those that require inference or interpretability in near-realtime. Thirdly, using activation maps sequences to identify useless layers/filters represents a novel approach towards network compression purposes. Fourthly, exploring the behavior of activation maps sequences for wrong classes and finding ways to exploit this information for classification purposes is a unique contribution. Lastly, coupling CAM-based videos with expert feedback in specific applications can result in a more interpretable and accurate model. Overall, these individual research contributions have the potential to improve the performance, efficiency, and interpretability of CNN models, leading to advancements in various image classification tasks and promoting large-scale and transparent adoption of these models.

\section*{Conclusion}
This work proposes CAManim as a novel XAI visualization method enabling end-users to better interpret CNN predictions by animating the CAM-based network activation maps through all layers. We demonstrate that CAManim works with any CAM-based method and various CNN architectures. We additionally introduce a quantitative end-to-end assessment inspired from the ROAD metric, denoted ``yellow brick ROAD" (ybROAD). Our experiments demonstrate the utility of these methods for improved interpretation and understanding of CNN predictions, not only in their final layers but across their layer-specific and global-wise perspectives. Visualizations and source code can be found at: \url{https://omni-ml.github.io/pytorch-grad-cam-anim/}

\section*{Acknowledgments}
The authors would like to acknowledge Dr. Katherine Muldoon for her support of this work. The authors also acknowledge that this study took place on unceded Algonquin Anishinabe territory.

\nolinenumbers

\bibliography{references} 

\end{document}


\vspace*{0.2in}

\begin{flushleft}
{\LARGE Supplementary Materials \\ \large
\textbf\newline{CAManim: Animating end-to-end network activation maps} 
}
\newline
\\
Emily Kaczmarek\textsuperscript{1,*},
Olivier X. Miguel\textsuperscript{2},
Alexa C. Bowie\textsuperscript{2},
Robin Ducharme\textsuperscript{2},
Alysha L.J. Dingwall-Harvey\textsuperscript{1,2},
Steven Hawken\textsuperscript{1,2,3,4},
Christine M. Armour\textsuperscript{1,5,6},
Mark C. Walker\textsuperscript{1,2,3,4,7,8,9,10},
Kevin Dick\textsuperscript{1,6,9*}
\\
\bigskip
\textbf{1} Children’s Hospital of Eastern Ontario Research Institute, Ottawa, Canada
\\
\textbf{2} Clinical Epidemiology Program, Ottawa Hospital Research Institute, Ottawa, Canada
\\
\textbf{3} School of Epidemiology and Public Health, University of Ottawa, Ottawa, Canada
\\
\textbf{4} ICES, Toronto, Canada
\\
\textbf{5} Department of Pediatrics, University of Ottawa, Ottawa, Canada
\\
\textbf{6} Prenatal Screening Ontario, Better Outcomes Registry \& Network, Ottawa, Canada
\\
\textbf{7} Department of Obstetrics and Gynecology, University of Ottawa, Ottawa, Canada
\\
\textbf{8} International and Global Health Office, University of Ottawa, Ottawa, Canada
\\
\textbf{9} BORN Ontario, Children’s Hospital of Eastern Ontario, Ottawa, Canada
\\
\textbf{10} Department of Obstetrics, Gynecology \& Newborn Care, The Ottawa Hospital, Ottawa, Canada
\\

\bigskip

\textcurrency Current Address: Children's Hospital of Eastern Ontario Research Institute, Ottawa, Ontario, Canada 
* \texttt{\{ekaczmarek, kdick\}@cheo.on.ca}

\end{flushleft}

\textbf{This document contains the supplementary materials in support of the main manuscript. Links to supporting code and datasets are made publicly available for use by the research community.}

\linenumbers

\section*{Code and Data Availability}

Publicly available examples of CAManim are available at: \href{https://omni-ml.github.io/pytorch-grad-cam-anim/intro.html}{https://omni-ml.github.io/pytorch-grad-cam-anim/intro.html}
\\~\\
The codebase implementing the CAManim method for animating end-to-end network activation maps is available at: \href{https://colab.research.google.com/github/OMNI-ML/pytorch-grad-cam-anim/blob/adapt-basecam-to-support-cam\_anim/tutorials/\_CAManim\_animating\_end2end\_activation\_maps.ipynb}{https://colab.research.google.com/github/OMNI-ML/pytorch-grad-cam-anim/blob/adapt-basecam-to-support-cam\_anim/tutorials/\_CAManim\_animating\_end2end\_activation\_maps.ipynb}
\\~\\
A Google Collab notebook demonstrating the use ov CAManim on the Mednist dataset using a DenseNet network is available at: \href{https://colab.research.google.com/github/OMNI-ML/pytorch-grad-cam-anim/blob/adapt-basecam-to-support-cam\_anim/tutorials/CAManim\_mednist\_tutorial.ipynb}{https://colab.research.google.com/github/OMNI-ML/pytorch-grad-cam-anim/blob/adapt-basecam-to-support-cam\_anim/tutorials/CAManim\_mednist\_tutorial.ipynb}